\documentclass[10pt,twocolumn,letterpaper]{article}

\usepackage{cvpr}
\usepackage{times}
\usepackage{epsfig}
\usepackage{graphicx}
\usepackage{amsmath}
\usepackage{amssymb}

\usepackage{mathtools}
\usepackage{booktabs}
\usepackage{amsfonts}       %
\usepackage{nicefrac}       %
\usepackage{microtype}      %
\usepackage{multirow}
\usepackage{multicol}
\usepackage{pifont}
\usepackage{soul,color}
\usepackage{subcaption}
\usepackage{afterpage}
\usepackage{tabularx, colortbl}
\usepackage{enumitem}

\DeclareMathOperator{\E}{\mathbb{E}}
\DeclareMathOperator{\KL}{\mathbb{KL}}
\DeclareMathOperator{\parallelbars}{%
  \,\|\,%
}
\renewcommand{\eqref}[1]{(\ref{#1})}

\newcommand\withparrallel[2]{#1\parallelbars#2}
\DeclarePairedDelimiter{\norm}{\lVert}{\rVert}

\DeclareMathOperator{\x}{\mathbf{x}}
\DeclareMathOperator{\xhat}{\hat{\x}}
\DeclareMathOperator{\z}{\mathbf{z}}
\DeclareMathOperator{\f}{\mathbf{f}}

\DeclareMathOperator{\zhat}{\hat{\z}}
\DeclareMathOperator{\zapp}{\z_\text{app}}
\DeclareMathOperator{\zlvis}{\z^\mathnormal{l}_\text{occ}}

\DeclareMathOperator{\zappi}{\z^\mathnormal{(i)}_\text{app}}
\DeclareMathOperator{\zlvisi}{\z^{\mathnormal{l(i)}}_\text{occ}}

\DeclareMathOperator{\zhatapp}{\hat{\z}_\text{app}}
\DeclareMathOperator{\zlhatvis}{\hat{\z}^\mathnormal{l}_\text{occ}}
\DeclareMathOperator{\zhatappi}{\hat{\z}^\mathnormal{(i)}_\text{app}}
\DeclareMathOperator{\zlhatvisi}{\hat{\z}^\mathnormal{l(i)}_\text{occ}}

\DeclareMathOperator{\zpriorvis}{\z^\text{prior}_\text{occ}}
\DeclareMathOperator{\zvis}{\z_\text{occ}}
\DeclareMathOperator{\zpart}{\z_\text{p}}
\DeclareMathOperator{\zparti}{\z^{\mathnormal{(i)}}_\text{p}}
\DeclareMathOperator{\zhatpart}{\hat{\z}_\text{p}}

\DeclareMathOperator{\Bern}{\operatorname{Bern}}
\DeclareMathOperator{\Norm}{\mathcal{N}}
\newcommand{\zivis}[1]{\z^\mathnormal{#1}_\text{occ}}
\newcommand{\zivisi}[1]{\z^\mathnormal{#1(i)}_\text{occ}}

\newcolumntype{C}[1]{>{\centering\arraybackslash\hspace{0pt}}p{#1}}

\usepackage[pagebackref=true,breaklinks=true,letterpaper=true,colorlinks,bookmarks=false]{hyperref}

\cvprfinalcopy %

\def\cvprPaperID{8893} %

\ifcvprfinal\fi
\begin{document}

\title{PatchVAE: Learning Local Latent Codes for Recognition}

\author{Kamal Gupta\textsuperscript{\rm 1}\\
\and
Saurabh Singh\textsuperscript{\rm 2} \\
\and
Abhinav Shrivastava\textsuperscript{\rm 1} \\
\and
\textsuperscript{\rm 1}University of Maryland, College Park\\
{\tt\small \{kampta,abhinav\}@cs.umd.edu}
\and
\textsuperscript{\rm 2}Google Research\\
{\tt\small saurabhsingh@google.com}
}

\maketitle

\begin{abstract}
Unsupervised representation learning holds the promise of exploiting large amounts of unlabeled data to learn general representations. A promising technique for unsupervised learning is the framework of Variational Auto-encoders (VAEs). However, unsupervised representations learned by VAEs are significantly outperformed by those learned by supervised learning for recognition. Our hypothesis is that to learn useful representations for recognition the model needs to be encouraged to learn about repeating and consistent patterns in data. Drawing inspiration from the mid-level representation discovery work, we propose PatchVAE, that reasons about images at patch level. Our key contribution is a bottleneck formulation that encourages mid-level style representations in the VAE framework. Our experiments demonstrate that representations learned by our method perform much better on the recognition tasks compared to those learned by vanilla VAEs.
\end{abstract}

\section{Introduction}
\label{sec:introduction}

Due to the availability of large labeled visual datasets, supervised learning has become the dominant paradigm for visual recognition. That is, to learn about any new concept, the modus operandi is to collect thousands of labeled examples for that concept and train a powerful classifier, such as a deep neural network. 
This is necessary because the current generation of models based on deep neural networks require large amounts of labeled data~\cite{Sun2017RevisitingUE}.
This is in stark contrast to the insights that we have from developmental psychology on how infants develop perception and cognition without any explicit supervision~\cite{smith2005development}. Moreover, the supervised learning paradigm is ill-suited for applications, such as health care and robotics, where annotated data is hard to obtain either due to privacy concerns or high cost of expert human annotators. In such cases, learning from very few labeled images or discovering underlying natural patterns in large amounts of unlabeled data can have a large number of potential applications. Discovering such patterns from unlabeled data is the standard setup of unsupervised learning.

\begin{figure}
    \centering
    \includegraphics[width=\columnwidth]{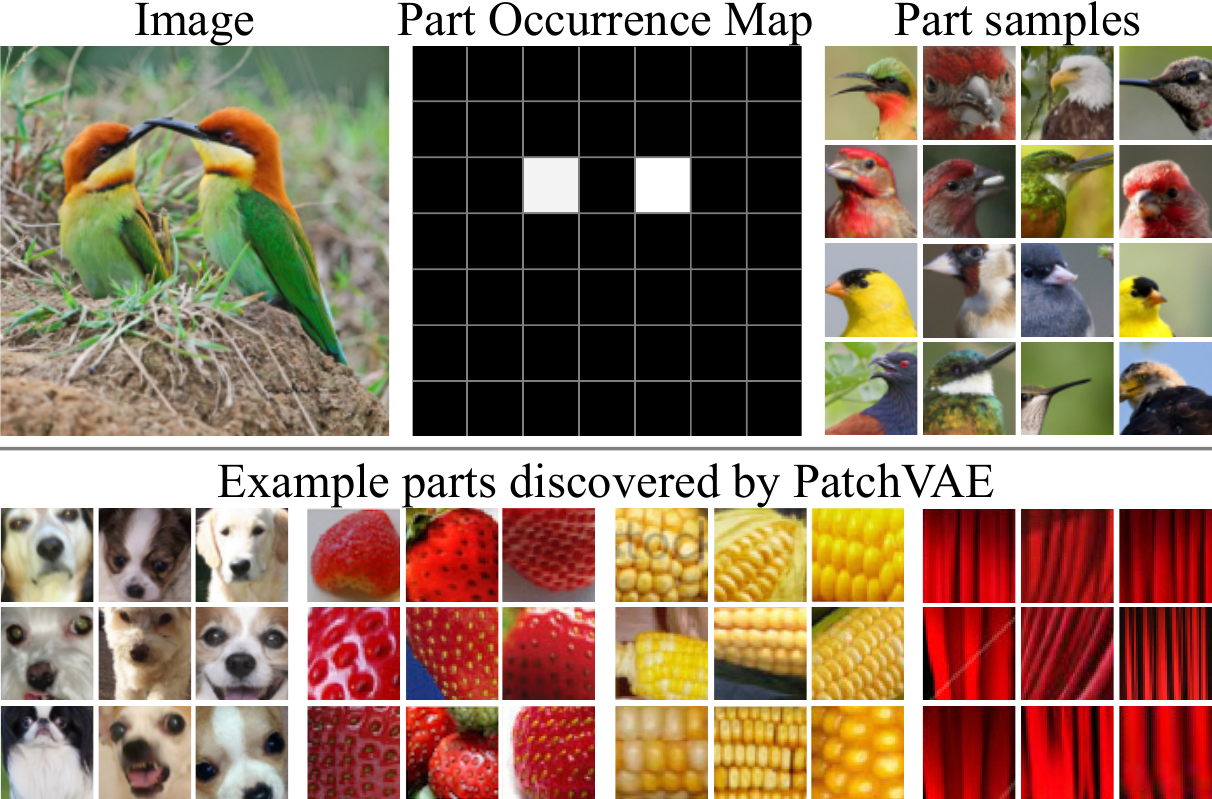}
    \vspace{-0.2in}
    \caption{PatchVAE learns to encode repetitive parts across a dataset, by modeling their appearance and occurrence. (top) Given an image, the occurrence map of a particular part learned by PatchVAE is shown in the middle, capturing the head/beak of the birds. Samples of the same part from other images are shown on the right, indicating consistent appearance. (bottom) More examples of parts discovered by our PatchVAE framework.}
    \label{fig:teaser}
    \vspace{-0.1in}
\end{figure}

Over the past few years, the field of unsupervised learning in computer vision has followed two seemingly different tracks with different goals: generative modeling and self-supervised learning. The goal of generative modeling is to learn the probability distribution from which data was generated, given some training data. Such models, learned using reconstruction-based losses, can draw samples from the same distribution or evaluate the likelihoods of new data, and are useful for learning compact representation of images. However, we argue that these representations are not as useful for visual recognition. This is not surprising since the task of reconstructing images does not require the bottleneck representation to sort out meaningful data useful for recognition and discard the rest; on the contrary, it encourages preserving as much information as possible for reconstruction.

In comparison, the goal in self-supervised learning is to learn representations that are useful for recognition. The standard paradigm is to establish proxy tasks that don't require human-supervision but can provide signals useful for recognition. Due to the mismatch in goals of unsupervised learning for visual recognition and the representations learned from generative modeling, self-supervised learning is a more popular way of learning representations from unlabeled data. However, fundamental limitation of this self-supervised paradigm is that we need to define a proxy-task that can mimic the desired recognition task. It is not possible to always establish such a task, nor are these tasks generalizable across recognition tasks.

In this paper, our goal is to enable the unsupervised generative modeling approach of VAEs to learn representations useful for recognition. Our key hypothesis is that for a representation to be useful, it should capture just the \emph{interesting} parts of the images, as opposed to \emph{everything} in the images.

What constitutes an interesting image part has been defined and studied in earlier works that pre-date the end-to-end trained deep network methods~\cite{Singh2012DiscPat,doersch2012what,juneja2013blocks}. Taking inspiration from these works, we propose a novel representation that only encodes few such parts of an image that are repetitive across the dataset, i.e., the patches that occur often in images. By avoiding reconstruction of the entire image our method can focus on regions that are repeating and consistent across many images. In an encoder-decoder based generative model, we constrain the encoder architecture to learn such repetitive parts -- both in terms of representations for appearance of these parts (or patches in an image) and where these parts occur. We formulate this using variational auto-encoder ($\beta$-VAEs)~\cite{kingma2013auto,Matthey2017betaVAELB}, where we impose novel structure on the latent representations. We use discrete latents to model part presence or absence and continuous latents to model their appearance. Figure~\ref{fig:teaser} shows an example of the discrete latents or occurrence map, and example parts discovered by our approach, PatchVAE. We present PatchVAE in Section~\ref{sec:approach} and demonstrate that it learns representations that are much better for recognition as compared to those learned by the standard $\beta$-VAEs~\cite{kingma2013auto,Matthey2017betaVAELB}. 

In addition, we present losses that favor foreground, which is more likely to contain repetitive patterns, in Section~\ref{subsec:betterloss}, and demonstrate that they result in representations that are much better at recognition. Finally, in Section~\ref{sec:quantitative}, we present results on CIFAR100~\cite{krizhevsky2009learning}, MIT Indoor Scene Recognition~\cite{quattoni2009recognizing}, Places~\cite{zhou2017places}, and ImageNet~\cite{deng2009imagenet} datasets. To summarize, our contributions are as follows:

\begin{enumerate}[leftmargin=*,noitemsep]
\item We propose a novel patch-based bottleneck in the VAE framework that learns representations that can encode repetitive parts across images.
\item We demonstrate that our method, PatchVAE, learns unsupervised representations that are better suited for recognition in comparison to traditional VAEs.
\item We show that losses that favor foreground are better for unsupervised representation learning for recognition.
\item We perform extensive ablation analysis of the proposed PatchVAE architecture.
\end{enumerate}

\vspace{-0.05in}
\section{Related Work}
\label{sec:related}
\vspace{-0.05in}
Due to its potential impact, unsupervised learning (particularly for deep networks) is one of the most researched topics in visual recognition over the past few years. Generative models such as VAEs~\cite{kingma2013auto,Matthey2017betaVAELB,kingma2016improved,gregor2015draw}, PixelRNN~\cite{DBLP:journals/corr/OordKK16}, PixelCNN~\cite{gulrajani2016pixelvae, Salimans2017PixeCNN}, and their variants have proven effective when it comes to learning compressed representation of images while being able to faithfully reconstruct them as well as draw samples from the data distribution. GANs~\cite{goodfellow2014generative, DBLP:journals/corr/RadfordMC15, zhu2017unpaired, arjovsky2017wasserstein} on the other hand, while don't model the probability density explicitly, can still produce high quality image samples from noise. There has been work combining VAEs and GANs to be able to simultaneously learn image data distribution while being able to generate high quality samples from it~\cite{khan2018adversarial, donahue2016adversarial, larsen2015autoencoding}. Convolution sparse coding~\cite{DBLP:journals/corr/abs-1804-02678} is an alternative approach for reconstruction or image in-painting problems. Our work complements existing generative frameworks in that we provide a structured approach for VAEs that can learn beyond low-level representations. We show the effectiveness of the representations learned by our model by using them for visual recognition tasks.

There has been a lot of work in interpreting or disentangling representations learned using generative models such as VAEs \cite{Matthey2017betaVAELB, Fraccaro2017ADR, kim2018disentangling}. However, there is little evidence of effectiveness of disentangled representations in visual recognition. Semi-supervised learning using generative models ~\cite{kingma2014semi, sohn2015learning}, where partial or noisy labels are available to the model during training, has shown lots of promise in applications of generating conditioned samples from the model. In our work however, we focus on incorporating inductive biases in these generative models (e.g., VAEs) so they can learn representations better suited for visual recognition. 

A related, but orthogonal, line of work is self-supervised learning where a proxy task is designed to learn representation useful for recognition. These proxy tasks vary from simple tasks like arranging patches in an image in the correct spatial order~\cite{doersch2014context, doersch2015unsupervised} and arranging frames from a video in correct temporal order~\cite{wang2015unsupervised, pathakCVPR17learning}, to more involved tasks like in-painting~\cite{pathak2016context} and context prediction~\cite{noroozi2016unsupervised, wang2017transitive}. We follow the best practices from this line of work for evaluating the learned representations.

\begin{figure*}
     \centering
     \begin{subfigure}[b]{0.17\textwidth}
         \centering
         \raisebox{3mm}{\includegraphics[width=\linewidth]{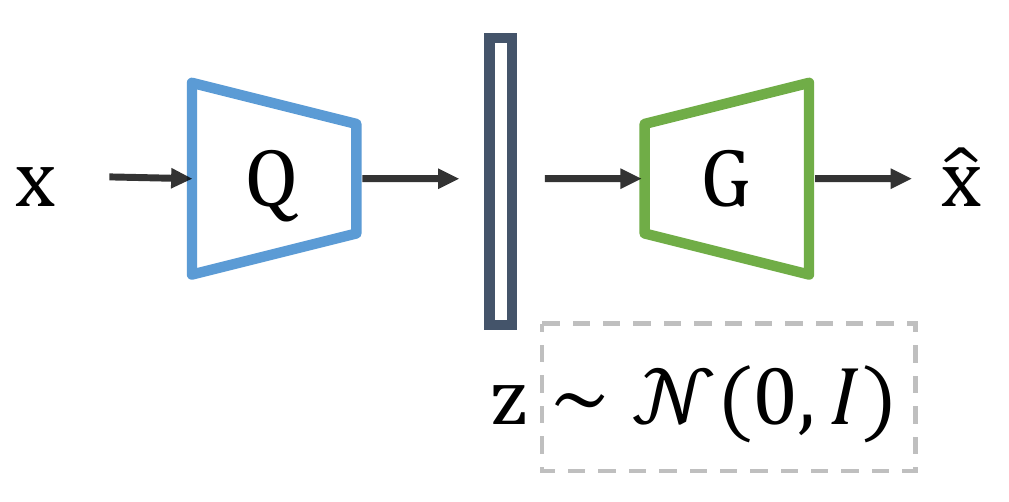}}
         \caption{\footnotesize VAE Architecture}
          \label{fig:vae}
     \end{subfigure}
     \quad\quad
     \begin{subfigure}[b]{0.75\textwidth}
         \centering
         \includegraphics[width=\linewidth]{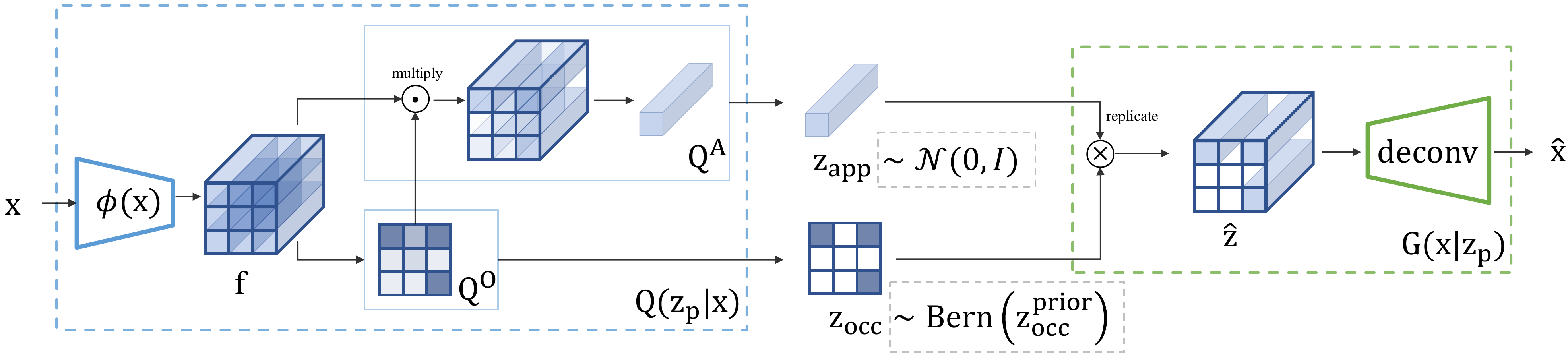}
         \caption{\footnotesize PatchVAE Architecture}
         \label{fig:overview}
     \end{subfigure}
     \vspace{-0.05in}
     \caption{(a) \textbf{VAE Architecture}: In a standard VAE architecture, output of encoder network is used to parameterize the variational posterior for $z$. Samples from this posterior are input to the decoder network. (b) \textbf{Proposed PatchVAE Architecture}: Our encoder network computes a set of feature maps $\f$ using $\phi(\x)$. This is followed by two independent single layer networks. The bottom network generates part occurrence parameters $Q^\text{O}$. We combine $Q^\text{O}$ with output of top network to generate part appearance parameters $Q^\text{A}$. We sample $\zvis$ and $\zapp$  to construct $\zhat$ as described in Section~\ref{subsec:patchyvae} which is input to the decoder network. We also visualize the corresponding priors for latents $\zapp$ and $\zvis$ in the dashed gray boxes.}
     \vspace{-0.15in}
\end{figure*}

\section{Our Approach}
\label{sec:approach}
Our work builds upon VAE framework proposed by~\cite{kingma2013auto}. We briefly review relevant aspects of the VAE framework and then present our approach. 

\subsection{VAE Review}
\label{subsec:vae}
Standard VAE framework assumes a generative model for data where first a latent $\mathbf{z}$ is sampled from a prior $p(\mathbf{z})$ and then the data is generated from a conditional distribution $G(\mathbf{x}|\mathbf{z})$. A variational approximation $Q(\mathbf{z}|\mathbf{x})$ to the true intractable posterior is introduced and the model is learned by minimizing the following negative variational lower bound (ELBO),
\begin{align}
\begin{split}
\mathcal{L}_\text{VAE}(\x) = 
&- \E_{z\sim Q(\mathbf{z}|\mathbf{x})}\left[ \log G(\mathbf{x}|\mathbf{z}) \right] \\[5pt]
&+ \KL\left[\withparrallel{Q(\mathbf{z}|\mathbf{x})}{p(\mathbf{z})} \right]
\end{split}
\label{eq:vae}
\end{align}
where $Q(\mathbf{z}|\mathbf{x})$ is often referred to as an encoder as it can be viewed as mapping data to the the latent space, while $G(\mathbf{x}|\mathbf{z})$ is referred to as a decoder (or generator) that can be viewed as mapping latents to the data space. Both $Q$ and $G$ are commonly parameterized as neural networks. Fig.~\ref{fig:vae} shows the commonly used VAE architecture.
If the conditional $G(\mathbf{x}|\mathbf{z})$ takes a gaussian form, negative log likelihood in the first term of RHS of Eq.~\ref{eq:vae} becomes mean squared error between generator output $\xhat=G(\mathbf{x}|\mathbf{z})$ and input data $\mathbf{x}$. In the second term, prior $p(\mathbf{z})$ is assumed to be a multi-variate normal distribution with zero-mean and identity covariance $\Norm\left(0,\,\mathcal{I}\right)$ and the loss simplifies to
\begin{align}
\begin{split}
\mathcal{L}_\text{VAE}(\x) = \norm{\mathbf{x}-\mathbf{\xhat}}^2 + \KL\left[\withparrallel{Q(\mathbf{z}|\mathbf{x})}{\Norm\left(0,\,\mathcal{I}\right)} \right]
\end{split}
\end{align}

When $G$ and $Q$ are differentiable, entire model can be trained with SGD using reparameterization trick \cite{kingma2013auto}. \cite{Matthey2017betaVAELB} propose an extension for learning disentangled representation by incorporating a weight factor $\beta$ for the KL Divergence term yielding
\begin{align}
\begin{split}
\mathcal{L}_{\beta\text{VAE}}(\x) = \norm{\mathbf{x}-\mathbf{\xhat}}^2 + \beta \KL\left[\withparrallel{Q(\mathbf{z}|\mathbf{x})}{\Norm\left(0,\,\mathcal{I}\right)} \right]
\end{split}
\label{eq:betavae}
\end{align}

VAE framework aims to learn a generative model for the images where the latents $\mathbf{z}$ represent the corresponding low dimensional generating factors. The latents $\mathbf{z}$ can therefore be treated as image representations that capture the necessary details about images. However, we postulate that representations produced by the standard VAE framework are not ideal for recognition as they are learned to capture \emph{all} details, rather than capturing `interesting' aspects of the data and dropping the rest. This is not surprising since there formulation does not encourage learning semantic information. For learning semantic representations, in the absence of any relevant supervision (as is available in self-supervised approaches), inductive biases have to be introduced. Therefore, taking inspiration from works on unsupervised mid-level pattern discovery~\cite{Singh2012DiscPat, doersch2012what,juneja2013blocks}, we propose a formulation that encourages the encoder to only encode such few parts of an image that are repetitive across the dataset, i.e., the patches that occur often in images. 

Since the VAE framework provides a principled way of learning a mapping from image to latent space, we consider it ideal for our proposed extension. We chose $\beta$-VAEs for their simplicity and widespread use. In Section~\ref{subsec:patchyvae}, we describe our approach in detail and in Section~\ref{subsec:betterloss} propose a modification in the reconstruction error computation to bias the error term towards foreground high-energy regions (similar to the biased initial sampling of patterns in~\cite{Singh2012DiscPat}).

\subsection{PatchVAE}
\label{subsec:patchyvae}

Given an image $\x$, let $\f = \phi(\x)$ be a deterministic mapping that produces a 3D representation $\f$ of size $h\times w\times d_e$, with a total of $L=h\times w$ locations (grid-cells).
We aim to encourage the encoder network to only encode parts of an image that correspond to highly repetitive patches.
For example, a random patch of noise is unlikely to occur frequently, whereas patterns like faces, wheels, windows, etc.\ repeat across multiple images. 
In order capture this intuition, we force the representation $\f$ to be useful for predicting frequently occurring parts in an image, and use \emph{just} these predicted parts to reconstruct the image. We achieve this by transforming $\f$ to  $\zhat$ which encodes a set of parts at a small subset of $L$ locations on the grid cells. We refer to $\zhat$ as ``patch latent codes'' for an image. Next we describe how we re-tool the $\beta$-VAE framework to learn these local latent codes. We first describe our setup for a single part and follow it up with a generalization to multiple parts (Section~\ref{sec:multipleparts}). 

\medskip
\noindent\textbf{Image Encoding.} Given the image representation $\f=\phi(x)$, we want to learn part representations at each grid location $l$ (where $l\in \{1,\dots,L\}$). A part is parameterized by its appearance $\zapp$ and its occurrence $\zlvis$ (i.e., presence or absence of the part at grid location $l$). We use two networks, $Q^\text{A}_{\f}$ and $Q^\text{O}_{\f}$, to parameterize posterior distributions $Q^\text{A}_{\f}(\zapp\,|\,\f)$ and $Q^\text{O}_{\f} (\zlvis\,|\,\f)$ of the part parameters $\zapp$ and $\zlvis$ respectively.
Since the mapping $\f=\phi(\x)$ is deterministic, we can re-write these distributions as $Q^\text{A}_{\f}(\zapp\,|\,\phi(\x))$ and $Q^\text{O}_{\f}(\zlvis\,|\,\phi(\x))$; or simply $Q^\text{A}(\zapp\,|\,\x)$ and $Q^\text{O}(\zlvis\,|\,\x)$. Therefore, given an image $\x$ the encoder networks estimate the posterior $Q^\text{A}(\zapp\,|\,\x)$ and $Q^\text{O}(\zlvis\,|\,\x)$. Note that $\f$ is a deterministic feature map, whereas $\zapp$ and $\zlvis$ are stochastic.

\medskip
\noindent\textbf{Image Decoding.} We utilize a generator or decoder network $G$, that given $\zvis$ and $\zapp$, reconstructs the image. First, we sample a part appearance $\zhatapp$ ($d_p$ dimensional, continuous) and then sample part occurrence $\zlhatvis$ ($L$ dimensional, binary) one for each location $l$ from the posteriors
\begin{align} 
\begin{split}
    \zhatapp &\sim Q^\text{A}(\zapp|\x)\\
\zlhatvis &\sim Q^\text{O}\left(\zlvis\,|\,\mathbf{x}\right), \;\;\; \text{where  } l\in \{1,\dots,L\}
\end{split}\label{eq:zhatsample}
\end{align}
Next, we construct a 3D representation $\zhat$ by placing $\zhatapp$ at every location $l$ where the part is present (i.e., $\zlhatvis=1$). This can be implemented by a broadcasted product of $\zhatapp$ and $\zlhatvis$ . We refer to $\zhat$ as \textbf{patch latent code}. Again note that $\f$ is deterministic and $\zhat$ is stochastic. Finally, a deconvolutional network takes $\zhat$ as input and generates an image $\xhat$. This image generation process can be written as 
\begin{align}
\xhat \sim G\left(\x \,|\,\zivis{1},\, \zivis{2},\, \dots,\, \zivis{L},\, \zapp \right)\label{eq:xhatsample}
\end{align}

Since all latent variables ($\zlvis$ for all $l$ and $\zapp$) are independent of each other, they can be stacked as
\begin{align}\label{eq:stack}
\zpart &= \left[\zivis{1};\, \zivis{2};\, \dots;\, \zivis{L};\, \zapp \right].
\end{align}
This enables us to use a simplified the notation (refer to \eqref{eq:zhatsample} and \eqref{eq:xhatsample}):
\vspace{-0.1in}\begin{align}
\begin{split}
\zhatpart &\sim Q^{\{\text{A,O}\}}(\zpart\,|\,\mathbf{x}) \\
\xhat &\sim G\left(\x \,|\,\zpart \right)
\end{split}\label{eq:simplesample}
\end{align}
Note that despite the additional structure, our model still resembles the setup of variational auto-encoders. The primary difference arises from: (1) use of discrete latents for part occurrence, (2) patch-based bottleneck imposing additional structure on latents, and (4) feature assembly for generator.

\medskip
\noindent\textbf{Training.} We use the training setup of $\beta$-VAE and use the maximization of variational lower bound to train the encoder and decoder jointly (described in Section~\ref{subsec:vae}). The posterior $Q^\text{A}$, which captures the appearance of a part, is assumed to be a Normal distribution with zero-mean and identity covariance $\Norm\left(0,\,\mathcal{I}\right)$. The posterior $Q^\text{O}$, which captures the presence or absence a part, is assumed to be a Bernoulli distribution $\Bern\left(\zpriorvis\right)$ with prior $\zpriorvis$. Therefore, the ELBO for our approach can written as (refer to~\eqref{eq:betavae}):
\begin{align}
\begin{split}
\mathcal{L_\text{PatchVAE}}(\x) = &-\E_{\zpart\sim Q^{\{\text{A,O}\}}(\zpart|\x)} \left[ G\left(\x \,|\,\zpart \right)\right] \\
&+\; \beta \KL\left[Q^{\{\text{A,O}\}}(\zpart|\x)\parallelbars p(\zpart)\right]
\end{split}\label{eq:losspatchy}
\end{align}
where, the $\KL$ term can be expanded as:
\begin{align}
\begin{split}
&\KL\left[Q^{\{\text{A,O}\}}(\zpart\,|\,\x)\parallelbars p(\zpart)\right] = \\
\beta_\text{app} \sum_{l=1}^L &\KL\left(Q^\text{O}(\zlvis\,|\,\mathbf{x}) \parallelbars \Bern\left(\zpriorvis\right)\right) \\
+\; \beta_\text{occ} &\KL\left(Q^\text{A}(\zapp\,|\,\mathbf{x}) \parallelbars \Norm\left(0,\mathcal{I}\;\right)\right)
\end{split}\label{eq:klpatchy}
\end{align}

\noindent\textbf{Implementation details.}
As discussed in Section~\ref{subsec:vae}, the first and second terms of the RHS of~\eqref{eq:losspatchy} can be trained using L2 reconstruction loss and reparameterization trick~\cite{kingma2013auto}. In addition, we also need to compute KL Divergence loss for part occurrence. Learning discrete probability distribution is a challenging task since there is no gradient defined to backpropagate reconstruction loss through the stochastic layer at decoder even when using the reparameterization trick. Therefore, we use the relaxed-bernoulli approximation~\cite{maddison2016concrete, agustsson2017soft} for training part occurrence distributions $\zlvis$. 

For an $H \times W$ image, network $Q(\f|\x)$ first generates feature maps of size $\left(h\times w\times d_e\right)$, where $(h,\,w)$ are spatial dimensions and $d_e$ is the number of channels. Therefore, the number of locations $L = h\times w$. Encoders $Q^\text{A}_{\f}(\zapp\,|\,\f)$ and $Q^\text{O}_{\f} (\zlvis\,|\,\f)$ are single layer neural networks to compute $\zapp$ and $\zlvis$. $\zlvis$ is $\left(h\times w\times 1\right)$-dimensional multivariate bernoulli parameter and $\zapp$ is $\left(1\times 1\times d_p\right)$-dimensional multivariate gaussian. $d_p$ is length of the latent vector for a single part. Input to the decoder $\zhat$ is $\left(h\times w\times d_p\right)$-dimensional. In all experiments, we fix $h=\frac{H}{8}$ and $w=\frac{W}{8}$.

\medskip
\noindent\textbf{Constructing $\zapp$.} Notice that $\f$ is an $\left(h\times w\times d_e\right)$-dimensional feature map and $\zlvis$ is $\left(h\times w\times 1\right)$-dimensional binary output, but $\zapp$ is $\left(1\times 1\times d_p\right)$-dimensional feature vector. If $\displaystyle \sum\nolimits_l \zlvis > 1$, the part occurs at multiple locations in an image. Since all these locations correspond to same part, their appearance should be the same. To incorporate this, we take the weighted average of the part appearance feature at each location, weighted by the probability that the part is present. Since we use the probability values for averaging the result is deterministic. This operation is encapsulated by the $Q^{\text{A}}$ encoder (refer to Figure~\ref{fig:overview}). During image generation, we sample $\zhatapp$ once and replicate it at each location where $\zlhatvis=1$. During training, this forces the model to: (1) only predict $\zlhatvis=1$ where similar looking parts occur, and (2) learn a common representation for the part that occurs at these locations. Note that $\zapp$ can be modeled as a mixture of distributions (e.g., mixture of gaussians) to capture complicated appearances. However, in this work we assume that the convolutional neural network based encoders are powerful enough to map variable appearance of semantic concepts to similar feature representations. Therefore, we restrict ourselves to a single gaussian distribution.

\subsection{PatchVAE with multiple parts}
\label{sec:multipleparts}
Next we extend the framework described above to use multiple parts. To use $N$ parts, we use $N\times2$ encoder networks $Q^{\text{A}(i)}\left(\zappi\,|\,\x\right)$ and  $Q^{\text{O}(i)}\left(\zlvisi\,|\,\x\right)$, where $\zappi$ and $\zlvisi$ parameterize the $i^\text{th}$ part. Again, this can be implemented efficiently as 2 networks by concatenating the outputs together. The image generator samples $\zhatappi$ and $\zlhatvisi$ from the outputs of these encoder networks and constructs $\zhat^{(i)}$. We obtain the final \textbf{patch latent code} $\zhat$ by concatenating all $\zhat^{(i)}$ in channel dimension. Therefore, $\zhat^{(i)}$ is $(h\times w\times d_p)$-dimensional and $\zhat$ is $(h\times w\times (N\times d_p))$-dimensional stochastic feature map. For this multiple part case,~\eqref{eq:stack} can be written as:
\begin{align}
\begin{split}
\z_\text{\textbf{P}} &= \left[\z^{(1)}_\text{p};\, \z^{(1)}_\text{p};\, \dots; \z^{(N)}_\text{p}\right]\\
\text{where  } \zparti &= \left[\zivisi{1};\, \zivisi{2};\, \dots;\, \zivisi{L};\, \zappi \right].
\end{split}\label{eq:multistack}
\end{align}
Similarly,~\eqref{eq:losspatchy} and~\eqref{eq:klpatchy} can be written as:
\begin{align}
\begin{split}
&\mathcal{L_\text{MultiPatchVAE}}(\x) = -\E_{\z_\text{\textbf{P}}} \left[ G\left(\x \,|\,\z_\text{\textbf{P}} \right)\right] \\
&+\; \beta_\text{app}\sum_{i=1}^{N} \sum_{l=1}^L \KL\left(Q^{\text{O}(i)}\left(\zlvisi\,|\,\mathbf{x}\right) \parallelbars \Bern\left(\zpriorvis\right)\right) \\
&+\; \beta_\text{occ}\sum_{i=1}^{N} \KL\left(Q^{\text{A}(i)}\left(\zappi\,|\,\mathbf{x}\right) \parallelbars \Norm\left(0,\mathcal{I}\;\right)\right)
\end{split}\label{eq:lossmultipatchy}
\end{align}
The training details and assumptions of posteriors follow the previous section.

\subsection{Improved Reconstruction Loss}
\label{subsec:betterloss}
The L2 reconstruction loss used for training $\beta$-VAEs (and other reconstruction based approaches) gives equal importance to each region of an image. This might be reasonable for tasks like image compression and image de-noising. However, for the purposes of learning semantic representations, not all regions are equally important. For example, ``sky'' and ``walls'' occupy large portions of an image, whereas concepts like ``windows,'' ``wheels,'', ``faces'' are comparatively smaller, but arguably more important. To incorporate this intuition, we use a simple and intuitive strategy to weigh the regions in an image in proportion to the gradient energy in the region. More concretely, we compute laplacian of an image to get the intensity of gradients per-pixel and average the gradient magnitudes in $8\times8$ local patches. The weight multiplier for the reconstruction loss of each $8\times8$ patch in the image is proportional to the average magnitude of the patch. All weights are normalized to sum to one. We refer to this as \textbf{weighted loss ($\mathcal{L}_\text{w}$)}. Note that this is similar to the gradient-energy biased sampling of mid-level patches used in~\cite{Singh2012DiscPat,doersch2012what}. Examples of weight masks are provided in the supplemental material.

In addition, we also consider an adversarial training strategy from GANs to train VAEs~\cite{larsen2015autoencoding}, where the discriminator network from GAN implicitly learns to compare images and gives a more abstract reconstruction error for the VAE. We refer to this variant by using `GAN' suffix in experiments. In Section~\ref{sec:qualitative}, we demonstrate that the proposed weighted loss ($\mathcal{L}_\text{w}$) is complementary to the discriminator loss from adversarial training, and these losses result in better recognition capabilities for both $\beta$-VAE and PatchVAE.

\section{Experiments}
\label{sec:quantitative}

\noindent\textbf{Datasets.}
We evaluate PatchVAE on CIFAR100~\cite{krizhevsky2009learning}, MIT Indoor Scene Recognition~\cite{quattoni2009recognizing}, Places~\cite{zhou2017places} and Imagenet~\cite{deng2009imagenet} datasets. CIFAR100 consists of 60k $32 \times 32$ color images from 100 classes, with 600 images per class. There are 50000 training images and 10000 test images. Indoor dataset contains 67 categories, and a total of 15620 images. Train and test subsets consist of 80 and 20 images per class respectively. Places dataset has 2.5 millions of images with 205 categories. Imagenet dataset has over a million images from 1000 categories.

\renewcommand{\arraystretch}{1.2}
\renewcommand{\tabcolsep}{2pt}
\begin{table*}[t]
    \centering
    \caption{Classification results on CIFAR100, Indoor67, and Places205. We initialize the classification model with the representations $\phi(\x)$ learned from unsupervised learning task. The model $\phi(\x)$ comprises of a conv layer followed by two residual blocks (each having 2 conv layers). First column (called `Conv1') corresponds to Top-1 classification accuracy with pre-trained model with the first conv layer frozen, second and third columns correspond to results with first three and first five conv layers frozen respectively. Details in Section~\ref{sec:results}\vspace{-0.05in}}
    \footnotesize{
    \begin{tabular}{@{}l|C{1.2cm}C{1.2cm}C{1.2cm}|C{1.2cm}C{1.2cm}C{1.2cm}|C{1.2cm}C{1.2cm}C{1.2cm}@{}}
    \toprule
    &   \multicolumn{3}{c|}{CIFAR100} & \multicolumn{3}{c|}{Indoor67} & \multicolumn{3}{c}{Places205}\\
    Model & Conv1 & Conv[1-3] & Conv[1-5] & Conv1 & Conv[1-3] & Conv[1-5] & Conv1 & Conv[1-3] & Conv[1-5] \\
    \midrule
    $\beta$-VAE & 44.12 & 39.65 & 28.57& 20.08 & 17.76 & 13.06& 28.29 & 24.34 & 8.89 \\
    $\beta$-VAE + $\mathcal{L}_\text{w}$ & 44.96& 40.30 & 28.33& 21.34 & 19.48 & 13.96& 29.43 & 24.93 & 9.41 \\
    $\beta$-VAE-GAN &44.69 & 40.13 & 29.89 &19.10 & 17.84 & 13.06& 28.48 & 24.51 & 9.72  \\
    $\beta$-VAE-GAN + $\mathcal{L}_\text{w}$ &45.61 & 41.35 & 31.53&20.45 & 18.36 & 14.33& 29.63 & 25.26 & 10.66 \\
    \midrule
    PatchVAE &43.07 & 38.58 & 28.72&20.97 & 19.18 & 13.43 & 28.63 & 24.95 & 11.09 \\
    PatchVAE + $\mathcal{L}_\text{w}$ &43.75 & 40.37 & 30.55& 23.21 & 21.87 & 15.45 & 29.39 & 26.29 & 12.07 \\
    PatchVAE-GAN &44.45 & 40.57 & 31.74& 21.12& 19.63 & 14.55& 28.87 & 25.25 & 12.21 \\
    PatchVAE-GAN + $\mathcal{L}_\text{w}$ &45.39 & 41.74 & 32.65 & 22.46 & 21.87 & 16.42 & 29.36 & 26.30 & 13.39  \\
    \midrule
    BiGAN & 47.72 & 41.89 & 31.58& 21.64 & 17.09&9.70 & 30.06 & 25.11 & 10.82  \\
    \midrule
    Imagenet Pretrained & 55.99 & 54.99 & 54.36 & 45.90 & 45.82 & 40.90 & 37.08 & 36.46 & 31.26\\
    \bottomrule
    \end{tabular}
    }
    \label{tab:benchmarks}
    \vspace{-0.07in}
\end{table*}

\begin{table}[t]
    \centering
         \caption{ImageNet classification results using ResNet18. We initialize weights from using the unsupervised task and fine-tune the last two residual blocks. Details in Section~\ref{sec:results}\vspace{-0.05in}}
    \renewcommand{\tabcolsep}{5pt}
    \renewcommand{\arraystretch}{1.2}
    \footnotesize{
    \begin{tabular}{@{}lcc@{}}
    \toprule
    Model & Top-1 Acc. & Top-5 Acc. \\
    \midrule
    $\beta$-VAE & 44.45 & 69.67\\
    PatchVAE & 47.01 & 71.71 \\
    \midrule
    $\beta$-VAE + $\mathcal{L}_\text{w}$ & 47.28 & 71.78 \\
    PatchVAE + $\mathcal{L}_\text{w}$ & 47.87 & 72.49\\
    \midrule
    Imagenet Supervised &  61.37 &  83.79 \\
    \bottomrule
    \end{tabular}
    }
    \label{tab:imagenet_benchmarks}
    \vspace{-0.1in}
\end{table}

\medskip
\noindent\textbf{Learning paradigm.} In order to evaluate the utility of PatchVAE features for recognition, we setup the learning paradigm as follows: we will first train the model in an unsupervised manner on all training images. After that, we discard the generator network and use only part of the encoder network $\phi(\x)$ to train a supervised model on the classification task of the respective dataset. We study different training strategies for the classification stage as discussed later.

\medskip
\noindent\textbf{Training details.} In all experiments, we use the following architectures. For CIFAR100, Indoor67, and Place205, $\phi(\x)$ has a conv layer followed by two residual blocks~\cite{he2016deep}. For ImageNet, $\phi(\x)$ is a ResNet18 model (a conv layer followed by four residual blocks). For all datasets, $Q^\text{A}$ and $Q^\text{O}$ have a single conv layer each. For classification, we start from $\phi(\x)$, and add a fully-connected layer with 512 hidden units and a final fully-connected layer as classifier. More details can be found in the supplemental material.

During the unsupervised learning phase of training, all methods are trained for 90 epochs for CIFAR100 and Indoor67, 2 epochs for Places205, and 30 epochs for ImageNet dataset. All methods use ADAM optimizer for training, with initial learning rate of $1\times10^{-4}$ and a minibatch size of 128. For relaxed bernoulli in $Q^\text{O}$, we start with the temperature of 1.0 with an annealing rate of $3\times10^{-5}$ (following the details in~\cite{agustsson2017soft}). For training the classifier, all methods use stochastic gradient descent (SGD) with momentum with a minibatch size of 128. Initial learning rate is $1\times10^{-2}$ and we reduce it by a factor of 10 every 30 epochs. All experiments are trained for 90 epochs for CIFAR100 and Indoor67, 5 epochs for Places205, and 30 epochs for ImageNet datasets.

\begin{figure}
    \centering
    \includegraphics[width=0.9\linewidth]{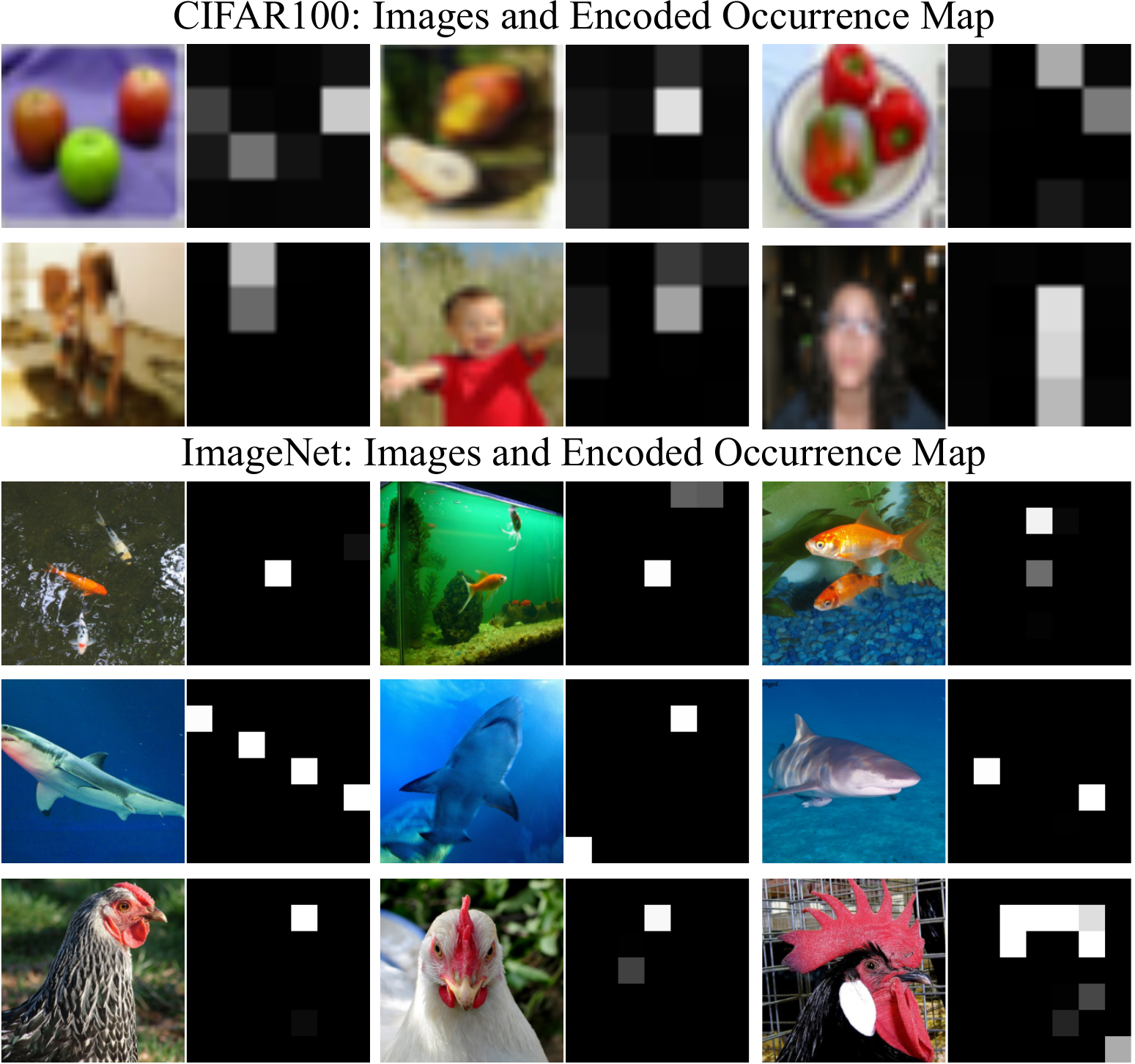}
    \vspace{-0.08in}
    \caption{Encoded part occurrence maps discovered on CIFAR100 and ImageNet. Each row represents a different part.}
    \vspace{-0.16in}
    \label{fig:vis}
\end{figure}

\medskip
\noindent\textbf{Baselines.} We use the \textbf{$\beta$-VAE} model (Section~\ref{subsec:vae}) as our primary baseline. In addition, we use weighted loss and discriminator loss resulting in the \textbf{$\beta$-VAE-*} family of baselines.
We also compare against a \textbf{BiGAN} model from~\cite{donahue2016adversarial}. We use similar backbone architectures for encoder/decoder (and discriminator if present) across all methods, and tried to keep the number of parameters in different approaches comparable to the best of our ability. Exact architecture details can be found in the supplemental material.

\begin{figure*}[t]
    \centering
    \includegraphics[width=\linewidth]{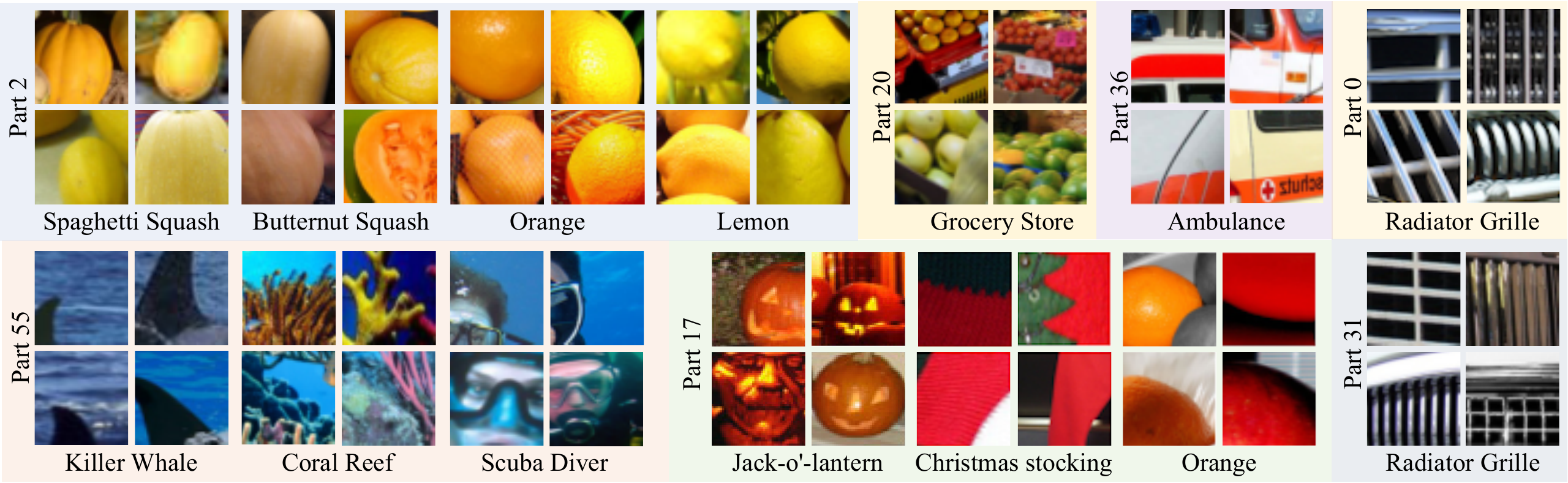}
    \vspace{-0.22in}
    \caption{A few representative examples for several parts to qualitatively demonstrate the visual concepts captured by PatchVAE. For each part, we crop image patches centered on the part location where it is predicted to be present. Selected patches are sorted by part occurrence probability as score. We manually select a diverse set from the top-50 occurrences from the training images. As can be seen, a single part may capture diverse set of concepts that are similar in shape or texture or occur in similar context, but belong to different categories. We show which categories the patches come from (note that category information was not used while training the model).}
    \label{fig:patches}
    \vspace{-0.15in}
\end{figure*}

\subsection{Downstream classification performance}
\label{sec:results}

In Table~\ref{tab:benchmarks}, we report the top-1 classification results on CIFAR100, Indoor67, and Places205 datasets for all methods with different training strategies for classification. First, we keep all the pre-trained weights in $\phi(\x)$ from the unsupervised task frozen and only train the two newly added conv layers in the classification network (reported under column `Conv[1-5]'). We notice that our method (with different losses) generally outperforms the $\beta$-VAE counterpart by a healthy margin. This shows that the representations learned by PatchVAE framework are better for recognition compared to $\beta$-VAEs. Moreover, better reconstruction losses (`GAN' and $\mathcal{L}_\text{w}$) generally improve both $\beta$-VAE and PatchVAE, and are complementary to each other.

Next, we fine-tune the last residual block along with the two conv layers (`Conv[1-3]' column). We observe that PatchVAE performs better than VAE under all settings except the for CIFAR100 with just L2 loss. However, when using better reconstruction losses, the performance of PatchVAE improves over $\beta$-VAE. Similarly, we fine-tune all but the first conv layer and report the results in `Conv1' column. Again, we notice similar trends, where our method generally performs better than $\beta$-VAE on Indoor67 and Places205 dataset, but $\beta$-VAE performs better CIFAR100 by a small margin. When compared to BiGAN, PatchVAE representations are better on all datasets (`Conv[1-5]') by a huge margin. However, when fine-tuning the pre-trained weights, BiGAN performs better on two out of four datasets. We also report results using pre-trained weights in $\phi(\x)$ using \emph{supervised} ImageNet classification task (last column, Table~\ref{tab:benchmarks}) for completeness. The results indicate that PatchVAE learns better semantic representations compared to $\beta$-VAE.

\medskip
\noindent\textbf{ImageNet Results.} Finally, we report results on the large-scale ImageNet benchmark in Table~\ref{tab:imagenet_benchmarks}. For these experiments, we use ResNet18~\cite{he2016deep} architecture for all methods. All weights are first learned using the unsupervised tasks. Then, we fine-tune the last two residual blocks and train the two newly added conv layers in the classification network (therefore, first conv layer and the following two residual blocks are frozen). We notice that PatchVAE framework outperforms $\beta$-VAE under all settings, and the proposed weighted loss helps both approaches. Finally, the last row in Table~\ref{tab:imagenet_benchmarks} reports classification results of same architecture randomly initialized and trained end-to-end on ImageNet using supervised training for comparison.

\begin{figure}[h]
    \centering
    \includegraphics[width=\linewidth]{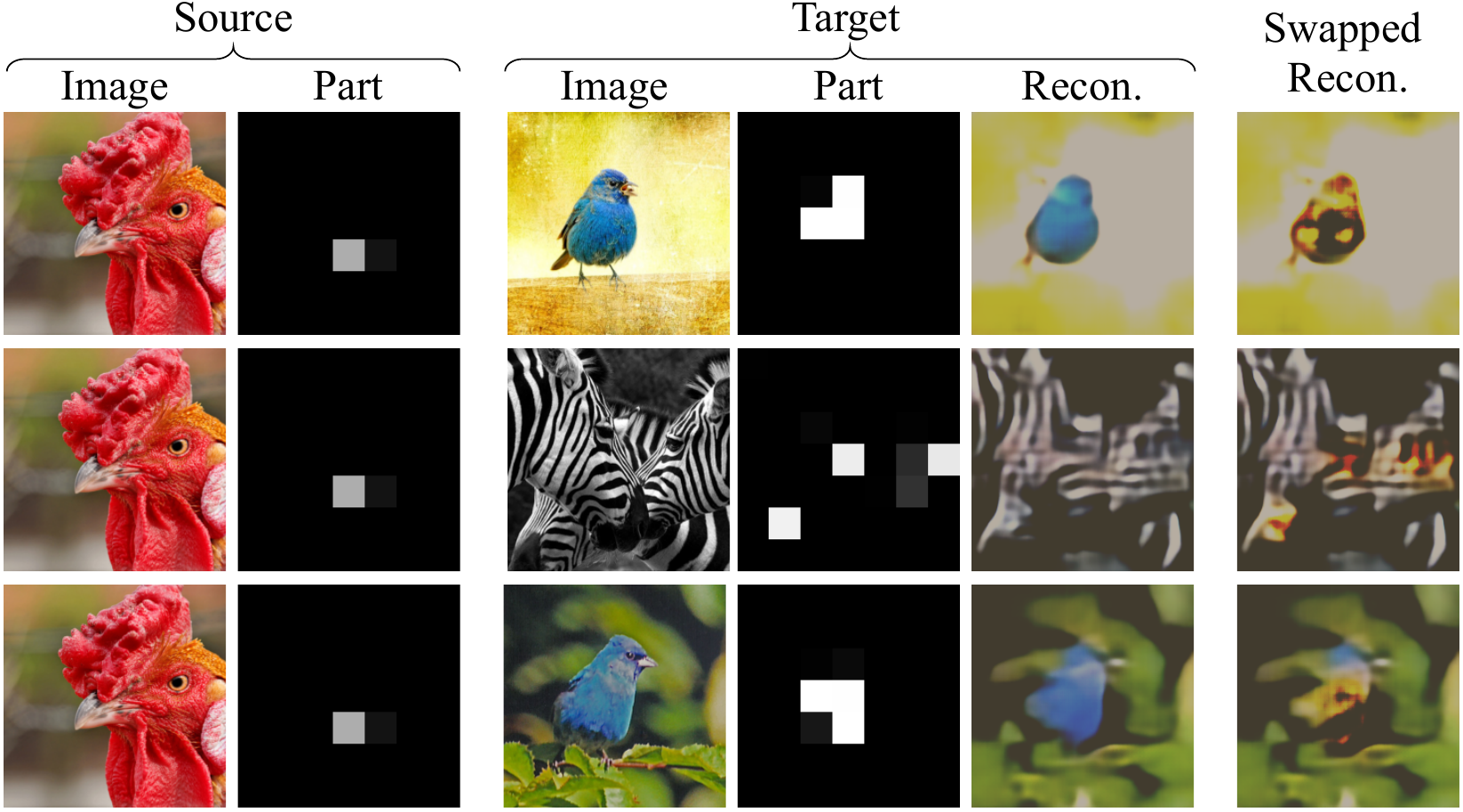}
    \vspace{-0.23in}
    \caption{Swapping source and target part appearance. Column 1, 2 show a source image with the occurrence map of one of the parts. We can swap the appearance vector of this part with appearance vectors of a different part in target images. Column 3, 4 show three target images with occurrence maps of one of their parts. Observe the change in reconstructions (column 5, 6) as we bring in the new appearance vector. The new reconstruction inherits properties of the source at specific locations in the target.}
    \vspace{-0.15in}
    \label{fig:partswap}
\end{figure}

\begin{table*}[t]
\centering
\parbox{0.26\linewidth}{
    \centering
    \caption{Effect of $N$: Increasing the maximum number of patches increases the discriminative power for CIFAR100 but has little or negative effect for Indoor67\vspace{-0.05in}}
    \footnotesize{
        \begin{tabular}{@{}ccc@{}}
        \toprule
        $N$ & CIFAR100 & Indoor67 \\
        \midrule
        4 & 27.59 & 14.40 \\
        8 & 28.74 & 12.69 \\
        16 & 28.94 & 14.33 \\
        32 & 27.78 & 13.28 \\
        64 & 29.00 & 12.76 \\
        \bottomrule
        \end{tabular}
    }
    \label{tab:np}
}\quad
\parbox{0.22\linewidth}{
    \centering
    \caption{Effect of $d_p$: Increasing the number of hidden units for a patch has very little impact on classification performance\vspace{-0.05in}}
    \footnotesize{
        \begin{tabular}{@{}ccc@{}}
        \toprule
        $d_p$ & CIFAR100 & Indoor67 \\
        \midrule
        3 & 28.63 & 14.25 \\
        6 & 28.97 & 14.55 \\
        9 & 28.21 & 14.55 \\
        \bottomrule
        \\
        \end{tabular}
    }
    \label{tab:nh}
}\quad
\parbox{0.2\linewidth}{
    \centering
    \caption{Effect of $\zpriorvis$: Increasing increasing the prior probability of patch occurrence has adverse effect on classification performance\vspace{-0.05in}}
    \footnotesize{
        \begin{tabular}{@{}ccc@{}}
        \toprule
        $\zpriorvis$ & CIFAR100 & Indoor67 \\
        \midrule
        0.01 & 28.86 & 14.33 \\
        0.05 & 28.67 & 14.25 \\
        0.1 & 28.31 & 14.03 \\
        \bottomrule
        \\
        \end{tabular}
    }
    \label{tab:py}
}\quad
\parbox{0.2\linewidth}{
    \centering
    \caption{Effect of $\beta_\text{occ}$: Too high or too low $\beta_\text{occ}$ can deteriorate the performance of learned representations\vspace{-0.05in}}
    \footnotesize{
        \begin{tabular}{@{}ccc@{}}
        \toprule
        $\beta_\text{occ}$ & CIFAR100 & Indoor67 \\
        \midrule
        0.06 & 30.11 & 14.10 \\
        0.3 & 30.37 & 15.67 \\
        0.6 & 28.90 & 13.51 \\
        \bottomrule
        \end{tabular}
    }
    \label{tab:betav}
}

\vspace{-0.05in}
\end{table*}

\begin{table}[t]
    \centering
    \footnotesize
    \caption{Reconstruction metrics on ImageNet. PatchVAE sacrifices reconstruction quality to learn discriminative parts, resulting in higher recognition performance (Table~\ref{tab:imagenet_benchmarks})\vspace{-0.05in}}
    \begin{tabular}{@{}lccc@{}}
    \toprule
    Model & PSNR $\uparrow$ & FID $\downarrow$ & SSIM $\uparrow$ \\
    \midrule
    $\beta$-VAE & 4.857 & 108.741 & 0.289 \\
    PatchVAE    & 4.342 & 113.692 & 0.235 \\
    \bottomrule
    \end{tabular}
    \vspace{-0.105in}
    \label{tab:gen}
\end{table}

\subsection{Qualitative Results}
\label{sec:qualitative}
We present qualitative results to validate our hypothesis. First, we visualize whether the structure we impose on the VAE bottleneck is able to capture occurrence and appearance of important parts of images. We visualize the PatchVAE trained on images from CIFAR100 and Imagenet datasets in the following ways.

\medskip
\noindent\textbf{Concepts captured.} First, we visualize the part occurrences in Figure~\ref{fig:vis}. We can see that the parts can capture round (fruit-like) shapes in the top row and faces in the second row regardless of the class of the image. Similarly for ImageNet, occurrence map of a specific part in images of chicken focuses on head and neck. Note that these semantically these parts are more informative than just texture or color what a $\beta$-VAE can capture. In Figure~\ref{fig:patches}, we show parts captured by the ImageNet model by cropping a part of image centered around the occurring part. We can see that parts are able to capture multiple concepts, similar in either shape, texture, or context in which they occur.

\medskip
\noindent\textbf{Swapping appearances.} Using PatchVAE, we can swap appearance of a part with the appearance vector of another part from a different image. In Figure~\ref{fig:partswap}, keeping the occurrence map same for a target image, we modify the appearance of a randomly chosen part and observe the change in reconstructed image. We notice that given the same source part, the decoder tries similar things across different target images. However, the reconstructions are worse since the decoder has never encountered this particular combination of part appearance before.

\medskip
\noindent\textbf{Discriminative \vs Generative strength.} As per our design, PatchVAE compromises the generative capabilities to learn more discriminative features. To quantify this, we use the the images reconstructed from $\beta$-VAE and PatchVAE models (trained on ImageNet) and compute three different metrics to measure the quality of reconstructions of test images. Table~\ref{tab:gen} shows that $\beta$-VAE is better at reconstruction.

\subsection{Ablation Studies}
\label{sec:ablation}

We study the impact of various hyper-parameters used in our experiments. For the purpose of this evaluation, we follow a similar approach as in the `Conv[1-5]' column of Table~\ref{tab:benchmarks} and all hyperparameters from the previous section. We use CIFAR100 and Indoor67 datasets for ablation analysis.

\medskip
\noindent\textbf{Maximum number of patches.} Maximum number of parts $N$ used in our framework. Depending on the dataset, higher value of $N$ can provide wider pool of patches to pick from. However, it can also make the unsupervised learning task harder, since in a minibatch of images, we might not get too many repeat patches. Table~\ref{tab:np}(left) shows the effect of $N$ on CIFAR100 and Indoor67 datasets. We observe that while increasing number of patches improves the discriminative power in case of CIFAR100, it has little or negative effect in case of Indoor67. A possible reason for this decline in performance for Indoor67 can be smaller size of the dataset (i.e., fewer images to learn).

\medskip
\noindent\textbf{Number of hidden units for a patch appearance $\zhatapp$.} Next, we study the impact of the number of channels in the appearance feature $\zhatapp$ for each patch ($d_p$). This parameter reflects the capacity of individual patch's latent representation. While this parameter impacts the reconstruction quality of images. We observed that it has little or no effect on the classification performance of the base features. Results are summarized in Table~\ref{tab:nh}(right) for both CIFAR100 and Indoor67 datasets.

\medskip
\noindent\textbf{Prior probability for patch occurrence $\zpriorvis$.}
In all our experiments, prior probability for a patch is fixed to $1/N$, i.e., inverse of maximum number of patches. The intuition is to encourage each location on occurrence maps to fire for at most one patch. Increasing this patch occurrence prior will allow all patches to fire at the same location. While this would make the reconstruction task easier, it will become harder for individual patches to capture anything meaningful. Table~\ref{tab:py} shows the deterioration of classification performance on increasing $\zpriorvis$.

\medskip
\noindent\textbf{Patch occurrence loss weight $\beta_\text{occ}$.} The weight for patch occurrence KL Divergence has to be chosen carefully. If $\beta_\text{occ}$ is too low, more patches can fire at same location and this harms the the learning capability of patches; and if $\beta_\text{occ}$ is too high, decoder will not receive any patches to reconstruct from and both reconstruction and classification will suffer. Table~\ref{tab:betav} summarizes the impact of varying $\beta_\text{occ}$.
\section{Conclusion}
We presented a patch-based bottleneck in the VAE framework that encourages learning useful representations for recognition. 
Our method, PatchVAE, constrains the encoder architecture to only learn patches that are repetitive and consistent in images as opposed to learning \emph{everything}, and therefore results in representations that perform much better for recognition tasks compared to vanilla VAEs. We also demonstrate that losses that favor high-energy foreground regions of an image are better for unsupervised learning of representations for recognition.

{\small
\bibliographystyle{ieee_fullname}
\bibliography{egbib}
}
\clearpage
\appendix

\section{Training Details}
\label{sec:app_datasets}
The generator network has two deconv layers with batchnorm and a final deconv layer with tanh activation. When training with `GAN' loss, the additional discriminator has four conv layers, two of which have batchnorm.

\section{Visualization of Weighted Loss}
\begin{figure*}[h]
    \centering
    \includegraphics[width=0.7\linewidth]{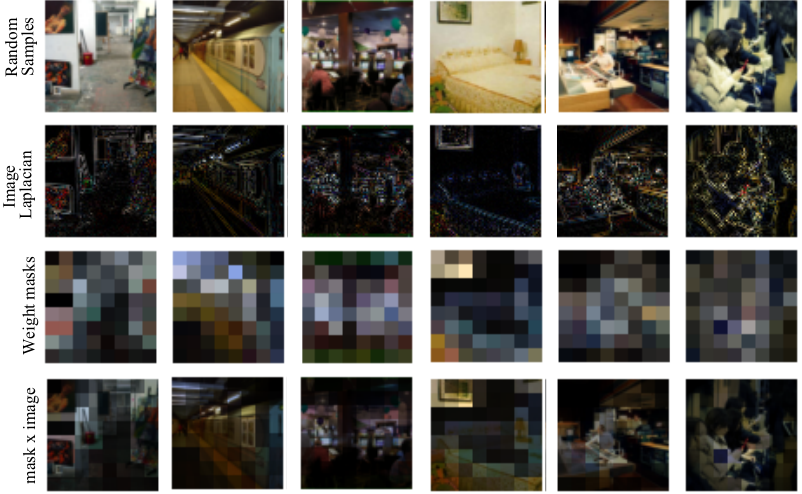}
    \caption{Masks used for weighted reconstruction loss $\mathcal{L}_\text{w}$. First row contains images randomly samples from MIT Indoor datatset. Second and third rows have the corresponding image laplacians and final reconstruction weight masks respectively. In the last row, we take the product of first and third row to highlight which parts of image are getting more attention while reconstruction.}
    \label{fig:masks}
\end{figure*}

Figure~\ref{fig:masks} shows an illustration of the reconstruction loss $\mathcal{L}_\text{w}$ proposed in Section 3.4. Notice that in first column, guitar has more weight that rest of the image. Similarly in second, fourth and sixth columns that train, painting, and people are respectively weighed more heavily by $\mathcal{L}_\text{w}$ than rest of the image; thus favoring capturing the foreground regions.

\section{Model Architecture}
In this section, we share the exact architectures used in various experiments. As discussed in Section 4, we evaluated our proposed model on CIFAR100, Indoor67, and Places205 datasets. We resize and center-crop the images such that input image size for CIFAR100 datasets is $32\times32\times3$ while for Indoor67 and Places205 datasets input image size is $64\times64\times3$. PatchVAE can treat images of various input sizes in exactly same way allowing us to keep the architecture same for different datasets. In case of VAE and BiGAN however, we have to go through a fixed size bottleneck layer and hence architectures need to be a little different for different input image sizes. Wherever possible, we have tried to keep the number of parameters in different architectures comparable.

\subsection{Architecture for unsupervised learning task}
Tables~\ref{tab:encoder32} and~\ref{tab:encoder64} show the architectures for encoders used in different models. In the unsupervised learning task, encoder comprises of a fixed neural network backbone $\phi(x)$, that given an image of size $h\times w\times3$ generated feature maps of size $\frac{h}{8}\times \frac{w}{8}\times d_e$. This backbone architecture is common to different models discussed in the paper and consists of a single conv layer followed by 2 residual blocks. We refer to this $\phi(x)$ as Resnet-9 and it is described as Conv1-5 layers in Table~\ref{tab:classifier}. Rest of the encoder architecture varies depending on the model in consideration and is described in the Tables~\ref{tab:encoder32} and~\ref{tab:encoder64}.

Tables~\ref{tab:decoder32} and~\ref{tab:decoder64} show the architectures for decoders used in different models. We use a pyramid like network for decoder where feature map size is doubled in consecutive layers, while number of channels is halved. Final non-linearity used in each decoder is tanh.

\subsection{Architecture for supervised learning task}
As discussed in Section 4, during the supervised learning phase, we discard rest of the encoder model and only keep $\phi(x)$ for classifier training. So the architectures for all baselines are exactly the same. Tables~\ref{tab:classifier} shows the architecture for classifier used in our experiments.

\begin{table*}[!htb]
    \centering
    \caption{Encoder architecture for unsupervised learning task on CIFAR100 - All `convolutional' layers are represented as $(\text{kernel}\_\text{size} \times \text{kernel}\_\text{size}, \text{channels}, \text{stride}, \text{pad})$. BN stands for batch normalization layer and ReLU for Rectified Linear Units.}
     \begin{tabular}{@{}l|ccc@{}}
    \toprule
    Layer & $\beta$-VAE & BiGAN & PatchVAE\\
    \midrule
    Features $\phi$ & Resnet-9 & Resnet-9 & Resnet-9 \\
    \midrule
    $ Q^\text{O} $ & - & - & $\begin{array} {lcl} (3\times3, 16, 1, 1) \end{array}$ \\
    \midrule
    $Q^\text{A}$ & 
    $\begin{array} {lcl}  (1\times1, 64, 1, 0) \\ \text{BN} \\ \text{ReLU}  \\ \mu: (4\times4, 96, 1, 0) \\ \sigma^2: (4\times4, 96, 1, 0) \end{array}$ & 
    $\begin{array} {lcl}  (1\times1, 64, 1, 0) \\ \text{BN} \\ \text{ReLU}  \\ (4\times4, 96, 1, 0) \end{array}$ & 
    $\begin{array} {lcl} \mu: (3\times3, 96, 1, 1) \\ \sigma^2: (3\times3, 96, 1, 1) \end{array}$ \\
    \midrule
    $\#$ Parameters & 888,192 & 789,792 & 922,896 \\
    \bottomrule
    \end{tabular}
    \label{tab:encoder32}
\end{table*}

\begin{table*}[!htb]
    \centering
    \caption{Encoder architecture for unsupervised learning task on Indoor67 and Places205 - All `convolutional' layers are represented as ($\text{kernel}\_\text{size} \times \text{kernel}\_\text{size}, \text{channels}, \text{stride}, \text{pad}$). BN stands for batch normalization layer and ReLU for Rectified Linear Units. Note that PatchVAE and $\beta$-VAE architectures are slightly different to account for sizes.}
    \begin{tabular}{@{}l|ccc@{}}
    \toprule
    Layer &$\beta$-VAE & BiGAN & PatchVAE \\
    \midrule
    Features $\phi$ & Resnet-9 & Resnet-9 & Resnet-9 \\
    \midrule
    $Q^\text{O}$ & - & - & $\begin{array} {lcl} (3\times3, 16, 1, 1) \end{array}$ \\
    \midrule
    $Q^\text{A}$ & $\begin{array} {lcl}  (1\times1, 64, 1, 0) \\ \text{BN} \\ \text{ReLU}  \\ \mu: (8\times8, 96, 1, 0) \\ \sigma^2: (8\times8, 96, 1, 0) \end{array}$ & $\begin{array} {lcl}  (1\times1, 64, 1, 0) \\ \text{BN} \\ \text{ReLU}  \\ (8\times8, 96, 1, 0) \end{array}$ & 
    $\begin{array} {lcl} \mu: (3\times3, 96, 1, 1) \\ \sigma^2:(3\times3, 96, 1, 1) \end{array}$ \\
    \midrule
    $\#$ Parameters & 1,478,016 & 1,084,704 & 922,896 \\
    \bottomrule
    \end{tabular}
    \label{tab:encoder64}
\end{table*}

\begin{table*}[!htb]
    \centering
    \caption{Decoder architecture for unsupervised earning task on CIFAR100 - All `deconvolutional' layers are represented as ($\text{kernel}\_\text{size} \times \text{kernel}\_\text{size}, \text{channels}, \text{stride}, \text{pad}$).  BN stands for batch normalization layer and ReLU for Rectified Linear Units.}
    \begin{tabular}{@{}l|ccc@{}}
    \toprule
     & $\beta$-VAE & BiGAN & PatchVAE \\
    \midrule
    Model & $\begin{array} {lcl} (4\times4, 64, 1, 0) \\ \text{BN} \\ \text{LeakyReLU}(0.2) \\ (1\times1, 256, 1, 0) \\ \text{BN} \\ \text{LeakyReLU}(0.2) \\ (4\times4, 128, 2, 1)  \\ \text{BN} \\ \text{LeakyReLU}(0.2) \\ (4\times4, 64, 2, 1)  \\ \text{BN} \\ \text{LeakyReLU}(0.2) \\ (4\times4, 3, 2, 1) \\ tanh \end{array}$ &
    $\begin{array} {lcl} (4\times4, 64, 1, 0) \\ \text{BN} \\ \text{LeakyReLU}(0.2) \\ (1\times1, 256, 1, 0) \\ \text{BN} \\ \text{LeakyReLU}(0.2) \\ (4\times4, 128, 2, 1)  \\ \text{BN} \\ \text{LeakyReLU}(0.2) \\ (4\times4, 64, 2, 1)  \\ \text{BN} \\ \text{LeakyReLU}(0.2) \\ (4\times4, 3, 2, 1) \\ tanh \end{array}$ & 
    $\begin{array} {lcl} (1\times1, 256, 1, 0) \\ \text{BN} \\ \text{LeakyReLU}(0.2) \\ (4\times4, 128, 2, 1)  \\ \text{BN} \\ \text{LeakyReLU}(0.2) \\ (4\times4, 64, 2, 1)  \\ \text{BN} \\ \text{LeakyReLU}(0.2) \\ (4\times4, 3, 2, 1) \\ tanh \end{array}$ \\
    \midrule
     $\#$ Parameters & 774,144 & 774,144 & 683,904 \\
    \bottomrule
    \end{tabular}
    \label{tab:decoder32}
\end{table*}

\begin{table*}[!htb]
    \centering
    \caption{Decoder architecture for unsupervised learning task on Indoor67 and Places205 - All `deconvolutional' layers are represented as ($\text{kernel}\_\text{size} \times \text{kernel}\_\text{size}, \text{channels}, \text{stride}, \text{pad}$). BN stands for batch normalization layer and ReLU for Rectified Linear Units. Note that PatchVAE and $\beta$-VAE architectures are slightly different to account for sizes.}
    \begin{tabular}{@{}l|ccc@{}}
    \toprule
    & $\beta$-VAE & BiGAN & PatchVAE \\
    \midrule
    Model & $\begin{array} {lcl} (8\times8, 64, 1, 0) \\ \text{BN} \\ \text{LeakyReLU}(0.2) \\ (1\times1, 256, 1, 0) \\ \text{BN} \\ \text{LeakyReLU}(0.2) \\ (4\times4, 128, 2, 1)  \\ \text{BN} \\ \text{LeakyReLU}(0.2) \\ (4\times4, 64, 2, 1)  \\ \text{BN} \\ \text{LeakyReLU}(0.2) \\ (4\times4, 3, 2, 1) \\ tanh \end{array}$ &
    $\begin{array} {lcl} (8\times8, 64, 1, 0) \\ \text{BN} \\ \text{LeakyReLU}(0.2) \\ (1\times1, 256, 1, 0) \\ \text{BN} \\ \text{LeakyReLU}(0.2) \\ (4\times4, 128, 2, 1)  \\ \text{BN} \\ \text{LeakyReLU}(0.2) \\ (4\times4, 64, 2, 1)  \\ \text{BN} \\ \text{LeakyReLU}(0.2) \\ (4\times4, 3, 2, 1) \\ tanh \end{array}$ & 
    $\begin{array} {lcl} (1\times1, 256, 1, 0) \\ \text{BN} \\ \text{LeakyReLU}(0.2) \\ (4\times4, 128, 2, 1)  \\ \text{BN} \\ \text{LeakyReLU}(0.2) \\ (4\times4, 64, 2, 1)  \\ \text{BN} \\ \text{LeakyReLU}(0.2) \\ (4\times4, 3, 2, 1) \\ tanh \end{array}$ \\
    \midrule
     $\#$ Parameters & 1,069,056 & 1,069,056 & 683,904 \\
    \bottomrule
    \end{tabular}
    \label{tab:decoder64}
\end{table*}

\begin{table*}[!htb]
    \centering
    \caption{Architecture for supervised learning task - same for all baselines and our model. All convolutional layers are represented as ($\text{kernel}\_\text{size} \times \text{kernel}\_\text{size}, \text{channels}, \text{stride}, \text{pad}$). BN stands for batch bormalization layer and ReLU for Rectified Linear Units. All pooling operations are MaxPool and are represented by ($\text{kernel}\_\text{size} \times \text{kernel}\_\text{size}, stride, pad$). Like Resnet-18, downsampling happens by convolutional layers that have a stride of 2. In our model, downsampling happens during Conv1, Pool, and after Conv4-5.}
    \begin{tabular}{@{}l|c|c|c@{}}
    \toprule
    Layer &  CIFAR100 ($32\times32\times3$) & Indoor67 ($64\times64\times3$) & Places205 ($64\times64\times3$) \\
    \midrule
    Conv1 & $1 \times \left\{\begin{array} {lcl} (7\times7, 64, 2, 3) \\ \text{BN} \\ \text{ReLU}  \\ \text{Pool}(3\times3, 2, 1) \end{array} \right.$ & $1 \times \left\{\begin{array} {lcl} (7\times7, 64, 2, 3) \\ \text{BN} \\ \text{ReLU}  \\ \text{Pool}(3\times3, 2, 1) \end{array} \right.$ & $1 \times \left\{\begin{array} {lcl} (7\times7, 64, 2, 3) \\ \text{BN} \\ \text{ReLU}  \\ \text{Pool}(3\times3, 2, 1) \end{array} \right.$ \\
    \midrule
    Conv2-3 & $2 \times \left\{\begin{array} {lcl} (3\times3, 64, 1, 1) \\ \text{BN} \\ \text{ReLU}  \\(3\times3, 64, 1, 1) \\ \text{BN} \end{array} \right.$  & $2 \times \left\{\begin{array} {lcl} (3\times3, 64, 1, 1) \\ \text{BN} \\ \text{ReLU}  \\(3\times3, 64, 1, 1) \\ \text{BN} \end{array} \right.$ & $2 \times \left\{\begin{array} {lcl} (3\times3, 64, 1, 1) \\ \text{BN} \\ \text{ReLU}  \\(3\times3, 64, 1, 1) \\ \text{BN} \end{array} \right.$ \\
    \midrule
    Conv4-5 & $2 \times \left\{\begin{array} {lcl} (3\times3, 128, 1, 1) \\ \text{BN} \\ \text{ReLU}  \\(3\times3, 128, 1, 1) \\ \text{BN} \end{array} \right.$  & $2 \times \left\{\begin{array} {lcl} (3\times3, 128, 1, 1) \\ \text{BN} \\ \text{ReLU}  \\(3\times3, 128, 1, 1) \\ \text{BN} \end{array} \right.$ & $2 \times \left\{\begin{array} {lcl} (3\times3, 128, 1, 1) \\ \text{BN} \\ \text{ReLU}  \\(3\times3, 128, 1, 1) \\ \text{BN} \end{array} \right.$ \\
    \midrule
    FC & $\begin{array}{lcl} 2048 \times 512 \\ 512 \times 100 \end{array}$ & $\begin{array}{lcl} 8192 \times 512 \\ 512 \times 67 \end{array}$ & $\begin{array}{lcl} 8192 \times 512 \\ 512 \times 205 \end{array}$ \\
    \midrule
    $\#$ Parameters & 1,783,460 & 4,912,259 & 4,983,053 \\
    \bottomrule
    \end{tabular}
    \label{tab:classifier}
\end{table*}

\vfill
\clearpage\vfill

\end{document}